\documentclass[arxiv]{melba}

\usepackage{amsmath,amsfonts}

\usepackage{url}
\usepackage{hyperref}
\usepackage{booktabs} 
\usepackage{multirow}
\usepackage[ruled,vlined]{algorithm2e}
\usepackage{pifont}
\usepackage[dvipsnames]{xcolor}
\newcommand{\cmark}{\color{green} \ding{51}}%
\newcommand{\xmark}{\color{red} \ding{55}}%

\melbaid{2025:021}  
\doi{10.59275/j.melba.2025-4582}
\melbaauthors{Galati*,Falcetta*, Cortese, Prados, Burgos, Zuluaga}  
\email{daniele.falcetta@eurecom.fr, maria.zuluaga@eurecom.fr}
\volume{3}
\firstpageno{477}  
\melbayear{2025}  
\datesubmitted{2025-03-03}  
\datepublished{2025-09-09}  

\ShortHeadings{Multi-Domain Brain Vessel Segmentation Through Feature Disentanglement}{Galati*,Falcetta*, Cortese, Prados, Burgos, Zuluaga}

\title{Multi-Domain Brain Vessel Segmentation Through Feature%
    \texorpdfstring{\\}{ }Disentanglement}

\author{
	\firstname Francesco \surname Galati*\aff{1}\orcid{0000-0001-6317-6298},
    \firstname Daniele \surname Falcetta*\aff{1}\orcid{0009-0009-7199-5424},
	\name Rosa Cortese\aff{2}\orcid{0000-0002-9803-7914},
    \name Ferran Prados\aff{3,4,5}\orcid{0000-0002-7872-0142},
    \name Ninon Burgos\aff{6}\orcid{0000-0002-4668-2006},
    \name Maria A. Zuluaga\aff{1,7}\orcid{0000-0002-1147-766X}
}
\affiliations{
	\num 1 \addr Data Science Department, EURECOM, Sophia Antipolis, FR \\
	\num 2 \addr Department of Medicine, Surgery and Neuroscience, University of Siena, IT \\
        \num 3 \addr e-Health Center, Universitat Oberta de Catalunya, Barcelona, ES \\
	\num 4 \addr Centre for Medical Image Computing (CMIC), Department of Medical Physics and Biomedical Engineering, University College London, UK \\
    \num 5 \addr Queen Square MS Centre, Department of Neuroinflammation, UCL Queen Square Institute of Neurology, Faculty of Brain Sciences, University College London, UK \\
    \num 6 \addr Sorbonne Université, Paris Brain Institute, CNRS, Inria, Inserm, AP-HP, Hôpital de la Pitié Salpêtrière, Paris, FR \\
    \num 7 \addr School of Biomedical Engineering \& Imaging Sciences, King's College London, UK\\
    \num * Joint first authors 
    }

\abstract{
The intricate morphology of brain vessels poses significant challenges for automatic segmentation models, which usually focus on a single imaging modality. However, accurately treating brain-related conditions requires a comprehensive understanding of the cerebrovascular tree, regardless of the specific acquisition procedure. Our framework effectively segments brain arteries and veins in various datasets through image-to-image translation while avoiding domain-specific model design and data harmonization between the source and the target domain. This is accomplished by employing disentanglement techniques to independently manipulate different image properties, allowing them to move from one domain to another in a label-preserving manner. Specifically, we focus on manipulating vessel appearances during adaptation while preserving spatial information, such as shapes and locations, which are crucial for correct segmentation. Our evaluation effectively bridges large and varied domain gaps across medical centers, image modalities, and vessel types. Additionally, we conduct ablation studies on the optimal number of required annotations and other architectural choices. The results highlight our framework's robustness and versatility, demonstrating the potential of domain adaptation methodologies to perform cerebrovascular image segmentation in multiple scenarios accurately.
	Our code is available at~\url{https://github.com/i-vesseg/MultiVesSeg}.}


\begin{document}

\twocolumn[\maketitle]

\section{Introduction}
\label{sec:introduction}

\begin{figure*}[t]
\centering
\includegraphics[width=\textwidth]{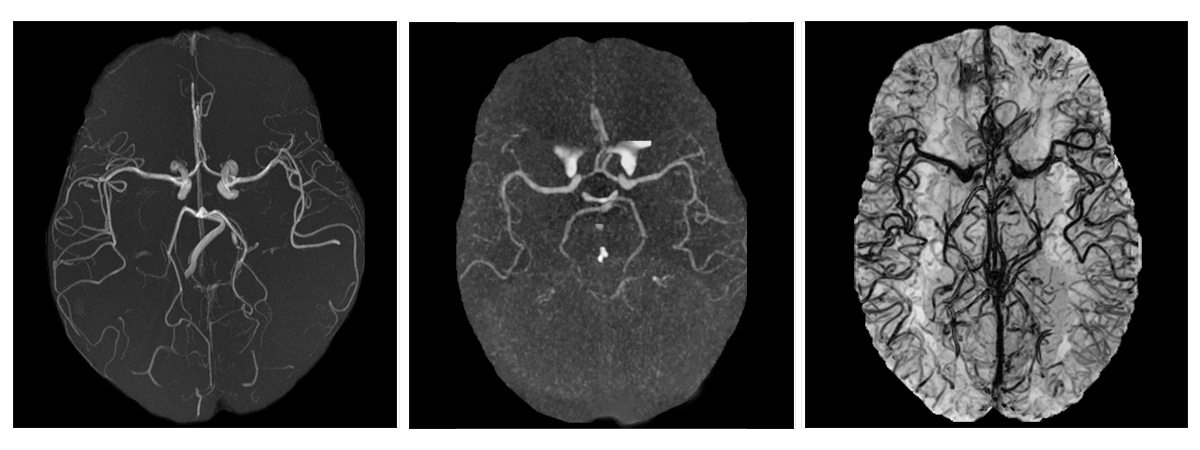}
\caption{Maximum intensity projection (MIP) of a magnetic resonance angiography (left), MIP of a computed tomography angiography (center), and minimum intensity projection (mIP) of a magnetic resonance venography (right). All images are skull-stripped and viewed from the axial perspective.} \label{Fig:ex_TOF_and_SWI}
\end{figure*}

Segmenting the cerebrovascular tree is crucial for accurately diagnosing and treating several brain-related conditions. The complex and intricate morphology of brain vessels requires the usage of multiple imaging modalities. Each modality has specific properties targeting a vessel type: angiographies focus on visualizing the arteries in the brain, while venographies primarily examine the veins. This variety of imaging modalities, combined with the different acquisition protocols and scanners utilized in clinical centers, poses challenges for automatic segmentation models, which struggle to generalize across different domains, i.e., varying centers, modalities, or vessel types (arteries or veins). When trained on a single source domain, models may become susceptible to distribution shifts, i.e., their performance may decline when transitioning from one domain to another. At the same time, developing, deploying, and maintaining a segmentation model for each domain is impractical, as collecting medical images is costly, and data annotation is laborious and demands a high level of expertise.

Distribution shift has been approached from various perspectives, depending on the amount of labeled data available. Among the different approaches, domain adaptation (DA) aims to transfer predictive knowledge from a source domain with abundant labeled data to a target domain with limited or no labeled data~\citep{Guan2022}. Despite numerous successful attempts to apply domain adaptation techniques in medical imaging, no current work has focused on brain vessel segmentation. Two factors may explain this. First, regardless of the number of modalities being examined, vessel segmentation remains challenging due to the relatively small size of vessels within a large image volume~\citep{Dang2022}, which can easily be merged with the background during adaptation~\citep{Tsai2018}. Second, the domain gap between existing modalities can vary widely.

The differences between the source and target data, referred to as domain-specific properties, can significantly impact the overall image appearance, encompassing various volume-related properties (e.g., spatial resolution, pixel spacing, image intensity range, contrast, or noise level). Moreover, differences can also manifest at a more localized level, impacting specific objects of interest: when examining vessel-related details in a brain scan, domain-specific variations can affect the vessels' intensities (e.g., vessels may be dark or bright), textures (e.g., vessels may have smooth or irregular surfaces), locations (e.g., vessels may be central or peripheral), shapes (e.g., vessels may be thick or thin), and densities (e.g., vessels may be more or less numerous).

Figure~\ref{Fig:ex_TOF_and_SWI} illustrates the visual disparity between a magnetic resonance angiography (MRA), a computed tomography angiography (CTA), and a magnetic resonance venography (MRV). This disparity can vary between different modalities. For example, arteries in MRA and CTA mainly differ in intensity distribution, as in the former, they stand out due to their high-intensity values, while in the latter, they blend with extracerebral tissue, making them harder to distinguish. Instead, the MRA-to-MRV domain gap also includes dissimilarities in the locations, shapes, and densities of the cerebral vasculature: although there is a correlation between the morphology of arteries and veins, the former are less numerous, occupy deeper positions within the brain tissue, and generally have larger sizes.

The larger the dissimilarities between source and target domains are, the more challenging it becomes to establish an image translation between the domains that facilitates effective segmentation. Indeed, translations from the target domain to the source cannot be performed fully, i.e., adapting all domain-specific properties to mimic the appearance of source images, as this would also involve vessel-related properties such as shapes, positions, and densities, which must be kept unchanged not to affect the final segmentation. To perform translations in a label-preserving way, i.e., generating hybrids between the source and the target domains, which only modify the necessary domain-specific properties to improve segmentation, we believe that additional disentangling mechanisms are required. To the best of our knowledge, these mechanisms are currently lacking, which would facilitate adapting vessels from diverse origins to a standard labeled source domain for segmentation.

In this work, we aim to develop a 3D brain vessel segmentation tool that can be applied out-of-the-box across any new target domain, avoiding domain-specific model design, which often demands significant expertise and is highly dependent on dataset properties~\citep{nnUnet}. To this end, we build on image-to-image translation and semantic segmentation techniques to formulate a semi-supervised domain adaptation framework that learns a \textit{disentangled representation} of image properties, allowing us to handle them independently.
\revision{The disentangled representation enables us to introduce a novel mechanism of \textit{label preservation}.} This mechanism identifies and translates only a subset of all the domain-specific features, discarding the ones that may compromise spatial information, such as vessel shapes and locations, which is crucial for correct segmentation.
As a result, our method eliminates the need for ad-hoc pre-processing steps commonly employed to homogenize image appearance across modalities, a process known in the literature as data harmonization~\citep{Eshaghzadeh2021}, thereby enhancing the flexibility of our model and simplifying domain adaptation across different domains. To assess model performance when enlarging the domain gap, we conduct evaluations in three increasingly complex scenarios: multi-center MRA, MRA-to-CTA, and MRA-to-MRV adaptation for vessel segmentation. Finally, we investigate the properties of the proposed framework through extensive ablation studies focusing on determining the optimal number of source and target annotations, assessing the efficacy of disentanglement, and testing different architectural choices that may impact the performance of our model.

\revision{While disentanglement techniques have been employed in broader computer vision contexts as noted by prior works such as DRANet~\citep{DRANet} and DRL-STNet~\citep{DRLSTNet}, our approach introduces novel contributions specifically tailored to cerebrovascular segmentation. Unlike previous methods that primarily focus on disentangling content and style for general anatomical structures, our framework uniquely separates vessel-specific properties (intensities, textures, shapes, locations, and densities) from volume-related properties. This fine-grained disentanglement is crucial for cerebrovascular structures due to their intricate morphology and the complexity of domain gaps between vessel imaging modalities.
Our method tackles the challenging problem of bridging the gap between fundamentally different vascular structures (arteries and veins) with distinct anatomical characteristics, advancing beyond existing image-to-image translation methods like XA-Sim2Real~\citep{XASim2Real} that typically address appearance transfer between similar vessel types. By leveraging path length regularization in a StyleGAN-based architecture~\citep{StyleGAN2}, we create a semantically organized latent space where different directions correspond to controllable aspects of variation, enabling independent manipulation of vessel intensities and textures while preserving spatial arrangements and geometrical properties.
Our approach is further distinguished by its two-phase training strategy that limits adversarial learning to only the initial phase, ensuring stable convergence when navigating extreme domain shifts. Additionally, our integrated label-synthesis branch enforces label-preserving translations focused on vessel geometries without requiring a separate segmentation module, reducing architectural design complexity. Our comprehensive evaluation across progressively wider domain gaps (multi-center MRA, MRA-to-CTA, and MRA-to-MRV) demonstrates the framework's effectiveness in scenarios that haven't been previously attempted in vessel segmentation domain adaptation.}

This paper is an extension of preliminary work~\citep{A2V}\revision{, significantly advancing our previous research through several key contributions.
\begin{itemize}
\item We have reformulated our framework to function as an out-of-the-box tool, eliminating the need for domain-specific design choices or ad-hoc data harmonization between source and target domains.
\item We introduce a \textit{label preservation mechanism} that ensures image-to-image translation preserves vessels shapes and locations when navigating wide domain gaps.
\item We extend our evaluation framework to include two additional clinically relevant scenarios: multi-center MRA adaptation and MRA-to-CTA adaptation, alongside our previous MRA-to-MRV scenario, allowing systematic analysis across progressively wider domain gaps.
\item We conduct new ablation studies investigating key aspects of our framework, including the impact of architectural choices, the minimum number of annotated samples required for effective adaptation, and the contribution of feature disentanglement to segmentation accuracy.
\end{itemize}}

\section{Related Work}
\label{sec:related_work}

\subsection{Multi-modal brain vessel segmentation}
The segmentation of the 3D cerebrovascular vessels has been widely explored in the literature~\citep{Chen2023}, encompassing different modalities and vessel types~\citep{Wilson1999,Beriault2015,Meijs2017}. Nonetheless, only a few works address multiple domains. In \cite{passat2007}, morphological operators simultaneously capture blood signals from paired time-of-flight MRA and T1-weighted MR sequences. In \cite{zuluaga2015}, a multi-scale tensor voting framework accounts for a voxel's scale and vicinity in paired CTA and 3D phase-contrast MR images. Despite developing a unified artery segmentation algorithm across image modalities, both studies~\citep{passat2007,zuluaga2015} require modality-specific initialization and parameter tuning. Recently, ~\cite{Tetteh2020} introduced an angiography segmentation model using 2D orthogonal cross-hair filters and a novel loss function for class imbalance with false-positive rate correction. After pre-training on synthetic data, the model is fine-tuned to segment human MRA data and CTA microscopy scans of rat brains, requiring pixel-wise annotations of each imaging modality. ~\cite{Dang2022} propose a weak patch-based deep learning approach for artery and vein segmentation from two MR sequences. A common limitation of these two methods is that they require separate training (or fine-tuning) for each domain. ~\cite{Chen2018} try to bypass domain-specific manual annotation by leveraging a paired dataset of MRA-CTA scans to generate annotations via registration, thresholding, and size filtering. However, the method faces limitations arising from misalignment between arteries after registration and the difficulty of acquiring paired datasets.

\subsection{Domain Adaptation}
Unsupervised domain adaptation (UDA) has been applied to the segmentation of various organs, such as liver~\citep{Hong2022}, lung~\citep{Li2022}, heart~\citep{SIFA,Wu2021}, abdominal structures~\citep{Hong2022Abdominal}, and brain substructures~\citep{SynthSeg}. The adaptation from the source to the target distribution can occur at different levels: input-level (or image-level)~\citep{Yao2022}, feature-level~\citep{Wu2022}, output-level~\citep{SynthSeg}, or as a combination of two of the previous categories~\citep{SIFA}. In recent years, unsupervised image-alignment methodologies have surged, driven by the advancements in neural style transfer~\citep{Gatys2015} and image-to-image translation~\citep{CycleGAN}, allowing for the extraction and combination of image content and style. Many image-alignment approaches, however, involve intricate architectures with multiple components~\citep{Ning2021} and heavily depend on adversarial training~\citep{SIFA}. Due to these factors, their behavior is known to be often unstable and difficult to interpret. To better guide the learning process, researchers have recently been redirecting their attention from fully unsupervised to semi-supervised DA for medical segmentation~\citep{Liu2022Semi,CS-CADA}, including limited target annotations into the training set.

In the specific context of DA for vessel segmentation, ~\cite{DCDA} leverages two segmentation models, each tailored to particular ophthalmic imaging modalities, operating within a UDA learning module to enhance the accuracy of 2D retinal vessel segmentation. ~\cite{CS-CADA} introduce a semi-supervised DA method designed for 2D cross-anatomy segmentation of 
coronary arteries and retinal vessels, integrating domain-specific batch normalization and cross-domain contrastive learning into a self-ensembling mean-teacher framework. Despite achieving promising results, the former technique might not be well-suited for large domain gaps due to the substantial disparity between brain arteries and veins. At the same time, the latter might face challenges due to the three-dimensional intricacies of the cerebrovascular structure.


\subsection{Domain Generalization}

The main limitation of DA is the requirement for repeated training with each novel target domain. Domain generalization enables a good performance across a wide range of target domains without the need to retrain~\citep{DGSurvey}. Among these, ~\cite{AADG} propose a data augmentation strategy for retinal vessel, optic disc and optic cup, and lesion segmentation that uses a model-agnostic augmentation policy to generate novel domains and maximize their diversity. ~\cite{Hu2024} introduce a novel domain generalization method integrating a Hessian-based vector field and self-attention mechanism to enhance tubular shape feature representation alongside a unique data augmentation preserving vessel structures while altering image style. Alternatively, foundation models like SAM~\citep{SAM} and SEEM~\citep{SEEM} have recently shown robust performance using vast training data and test-time prompts such as points, bounding boxes, masks, or text that guide the segmentation tasks. However, their practical use is hindered by the need for fine-tuning, interactive filtering of extraneous predictions, and high computational costs~\citep{Ma2023,Huang2023}. Thus, they are practically hard to adapt when annotations are rare, and datasets are small, such as in biomedical applications. In contrast, in-context learning methods adapt to new tasks without additional training by incorporating task demonstrations as inputs. Among these, UniverSeg~\citep{UniverSeg} has shown promise in medical image segmentation tasks by prompting support sets of image-label pairs, outperforming few-shot baseline methods. Although UniverSeg considers retinal vessels and has demonstrated its capability to generalize to unseen anatomies, we argue that requiring UniverSeg to bridge large domain gaps, such as the one from retinal to brain vessels, without incorporating any adaptation mechanism might be highly demanding. 

\section{Method}
\label{sec:method}

\begin{figure*}[ht]
\centering
\includegraphics[width=\textwidth]{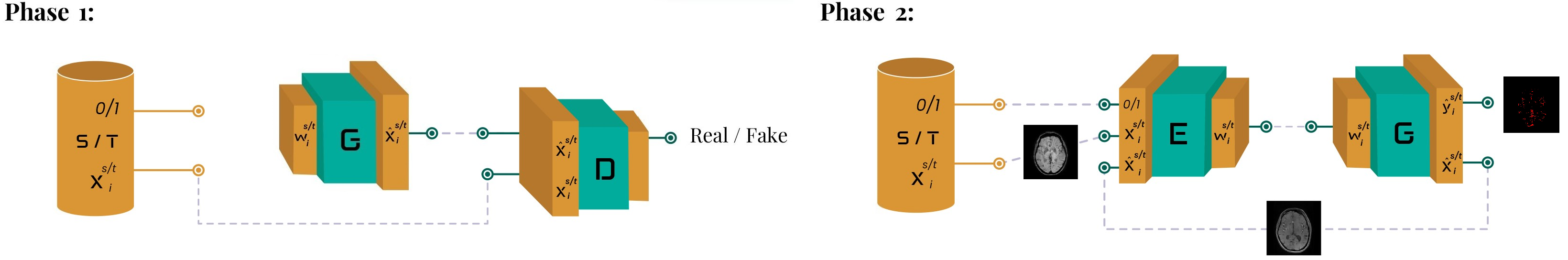}
\caption{During the two-phase training algorithm, images $x_i$ from domains $\mathcal{S}$ and $\mathcal{T}$ are input into our model consisting of the generator $G$, discriminator $D$, and encoder $E$. The training process is split into two distinct phases. In Phase 1 (left), $G$ undergoes adversarial training with $D$ to build a unified latent space that is both disentangled and semantically rich. In Phase 2 (right), the encoder $E$ is trained for label-preserving image-to-image translation, while $G$ is refined to generate segmentation masks $\hat{y}_i^t$ and $\hat{y}_i^s$.} \label{Fig:sketch}
\end{figure*}

Let $\mathcal{S}$ represent the source domain, and $\mathcal{T}$ represent a target domain. Our framework relies on three datasets: $S = \{x_i^s, y_i^s\}_{i=1}^N$, a set of $N$ labeled images from $\mathcal{S}$; $T_{U} = \{x_i^t\}_{i=1}^M$, a set of $M$ unlabeled images from $\mathcal{T}$; and $T_{L} = \{x_i^{lt}, y_i^{lt}\}_{i=1}^m$, a target labeled dataset with $m \ll M$ annotated images, also from $\mathcal{T}$. We denote $T = T_{U} \cup \{x_i^{lt} \mid x_i^{lt} \in T_{L}\}_{i=1}^m$ as the set of all target samples, excluding the labels $y_i^{lt}$. 

Our framework comprises a generator ($G$) and an encoder ($E$) accomplishing distinct tasks (Figure~\ref{Fig:sketch}). The generator learns to generate realistic brain images, $\hat{x}$, by identifying the features from the source and target domains and representing them both within a unified and \textit{disentangled} latent space $\mathcal{W}$. 
This representation allows our model to independently manipulate domain-specific features, enabling it to bridge broad domain gaps and compensate for the absence of data harmonization between the source and target at pre-processing. The encoder leverages the information in $\mathcal{W}$ to learn image-to-image translation in a \textit{label-preserving} manner, i.e., focusing only on features that do not compromise spatial information. This is achieved using \textit{cycle-consistency} and segmentation losses that enforce $E$ to maintain the labels aligned in both domains.

\revision{Both $G$ and $E$ are trained in separate phases. This two-phase training strategy represents a crucial architectural decision that significantly impacts both training stability and segmentation performance. Splitting the training process into two distinct phases limits the adversarial training solely to the first phase, where an external discriminator $D$ is incorporated to distinguish between real and fake images. By isolating adversarial training to Phase 1, we avoid the well-known instability issues associated with continuous adversarial learning throughout the entire adaptation process. The discriminator's role is deliberately confined to the initial phase, where it helps establish a disentangled latent space capable of representing both domains.

Once this foundation is established, removing the discriminator in Phase 2 allows the model to freely generate hybrid translations between domains without being penalized for producing outputs that do not strictly adhere to either domain's distribution. Excluding $D$ from the second phase, when the network learns image-to-image translation, prevents penalization of hybrid translations. This is particularly important when performing cross-domain translation between arteries and veins, which may require intermediate representations that would otherwise be rejected by a discriminator trained to distinguish between pure domain samples. Furthermore, this approach reduces \revision{architectural design} complexity and prevents the adversarial component from interfering with the label-preserving mechanisms that operate during the image-to-image translation in Phase 2.}

\subsection{Feature Disentanglement}
In Phase 1 (Figure~\ref{Fig:sketch} left), $G$ is trained to establish an association between latent vectors $w$ randomly sampled from $\mathcal{W}$ and the corresponding generated brain images, $\hat{x}$, which aim to resemble images from $\mathcal{S}$ or $\mathcal{T}$. To this end, we rely on adversarial learning with the aid of an external discriminator $D$. $D$ acts as a binary classifier distinguishing between real and fake samples. In response, $G$ aims to fool the discriminator by retrieving images that mimic the original ones from $S$ and $T$. The parameters of $G$ and $D$ are optimized with the following loss function:

\begin{equation}
\label{eq:phase_1}
\mathcal{L}_{tot} = \mathcal{L}_{adv}(G,D) + \mathcal{L}_{R_1}(D) + \revision{\mathcal{L}_{pl}(G)},
\end{equation}
where $\mathcal{L}_{adv}$ is the non-saturating loss~\citep{nonSaturatingLoss}, $\mathcal{L}_{R_1}$ is the $R_1$ regularization~\cite{R1regularization}, and $\mathcal{L}_{pl}$ is the path length regularization~\citep{StyleGAN2}.

\revision{The regularization brought by $\mathcal{L}_{\mathrm{pl}}$ transforms $\mathcal{W}$ into a disentangled latent space where different directions consistently correspond to individual, controllable aspects of variation in the generated images. This regularization encourages smooth transitions in the generated outputs when traversing the latent space, penalizing rapid changes by calculating $\mathcal{L}_{\mathrm{pl}} = \mathbb{E}[(\|\nabla_w G(w)\|_2 - a)^2]$, where $a$ is a running average of the gradient norm computed over previous iterations. As a result, the latent space becomes more uniform and semantically organized, facilitating the independent manipulation of domain-specific features.}

At the end of Phase 1, $\mathcal{W}$ can be queried to summarize the characteristics of both $\mathcal{S}$ and $\mathcal{T}$ in a shared and unwarped representation. Accordingly, this representation integrates the distinctive features from each domain, i.e., the domain-specific features.

In Phase 2 (Figure~\ref{Fig:sketch} right), $E$ is trained. When fed with an image $x_i^{s}$ from $S$, $E$ learns to discover two corresponding latent representations, i.e., $w_i^{s}$ and $w_i^{t}$, which are alternated by inputting an additional binary flag $d$, \revision{which functions as a domain selector that conditions the encoder's behavior: when $d=0$, the encoder maps the input to the source domain's region in the latent space $\mathcal{W}$, and when $d=1$, it maps to the target domain's region. This mechanism enables the model to perform both intra-domain reconstruction and cross-domain translation using a single encoder architecture, significantly reducing architectural design complexity compared to approaches that use separate encoders for each domain.}
The latent vectors guide $G$, which in this phase acts as a static decoder (with frozen parameters), to retrieve the source reconstruction $\hat{x}_i^{s}$, within the same domain. 
\begin{equation}
\label{eq:intra-domain}
\hat{x}_i^{s}=G(w_i^{s})=G(E(x_i^{s} \mid d=0)),
\end{equation}
or the source-to-target translation $\hat{x}_i^{t}$ to the opposite domain, 
\begin{equation}
\label{eq:inter-domain}
\hat{x}_i^{t}=G(w_i^{t})=G(E(x_i^{s} \mid d=1)).
\end{equation}

When learning $w_i^{s}$ and $w_i^{t}$, $E$ must encode the domain-specific features that recall the characteristics of either the source or target domain. \textit{Disentanglement} ensures that all image properties, whether related to the whole volume (e.g., pixel spacing or image contrast) or specific to vessels (e.g., intensities, textures, shapes, locations, and densities of vessels), are individually represented within $\mathcal{W}$, facilitating $E$ in establishing mappings between images at flexible semantic levels.


\subsection{\revision{Label Preservation Mechanism}}
\revision{In the specific context of brain vessels, domain-specific and domain-invariant features may vary depending on the source and targeted modalities. For example, images from two different angiographic techniques (e.g., Time-of-Flight MRA and CTA) will share a set of domain-specific and invariant features, as they both focus on the brain arteries, but this will not hold if the target domain is a venography, since vein geometry and location differ largely from that of the arteries. In this scenario, conventional approaches tend to modify domain-specific features indiscriminately, including those related to vessel geometry and positioning. Altering these attributes compromises segmentation accuracy, as the resulting translations no longer align with the original annotations. 

To address this limitation, we introduce a new label preservation mechanism, a crucial component of our framework, which ensures information integrity during domain adaptation. This mechanism enables the learning of domain-invariant and domain-specific features across different domains by enforcing consistency between segmentation masks before and after translation. More specifically, }
%
%
in Phase 2, we integrate image segmentation into our framework by expanding the generator with an additional label-synthesis branch $G_{lsb}$~\citep{DatasetGAN} (Figure~\ref{Fig:sketch} right). This branch is designed to output semantic segmentation masks that align with the generated images: while $G$ renders the source reconstruction $\hat{x}_i^{s}$ and the source-to-target translation $\hat{x}_i^{t}$, its label-synthesis branch predicts the associated segmentation maps $\hat{y}_i^{s}$ and $\hat{y}_i^{t}$. With this branch, we avoid using a separate segmentation module, thus decreasing \revision{architectural design} complexity. It consists of three fully connected layers attached to the feature vectors of $G$, which are optimized in isolation while freezing all the other parameters inside the generator.
\revision{To carry out this optimization, segmentation losses~$\mathcal{L}_{\textrm{\textsc{s}}}$ are computed for both $\hat{y}_i^{s}$ and $\hat{y}_i^{t}$ based on the same reference annotation $y_i^{s}$. The segmentation loss is calculated as:

\begin{equation}
\mathcal{L}_{\textrm{\textsc{s}}} = \mathcal{L}_{\textrm{dice}} + \mathcal{L}_{\textrm{ce}},
\end{equation}
where $\mathcal{L}_{\textrm{ce}}$ is the cross-entropy loss, and $\mathcal{L}_{\textrm{dice}}$ is the Dice loss defined as:

\begin{equation}
\mathcal{L}_{\textrm{dice}} = -\frac{2 \cdot \textrm{TP} + \epsilon}{2 \cdot \textrm{TP} + \textrm{FP} + \textrm{FN} + \epsilon},
\end{equation}
where TP, FP, and FN represent true positives, false positives, and false negatives respectively, and $\epsilon=10^{-5}$ is a small constant to prevent division by zero.}

Requiring the model to output the same segmentation masks post-reconstruction and post-translation is crucial to guarantee that labels are preserved during both processes. This requirement backpropagates to $E$ and ensures that $w_i^{s}$ and $w_i^{t}$ share the necessary domain-specific features to preserve the position and shapes of objects, particularly vessels (as it is our object of interest), which are consequently excluded from the translation process. 
For example, transforming pixel spacing (i.e., a domain-specific feature) in case it differs between $\mathcal{S}$ and $\mathcal{T}$ may increase or decrease the overall image scale, thus negatively affecting the segmentation. For this reason, $E$ will avoid translating pixel spacing. 
Here, disentanglement proves beneficial, helping the model separate domain-specific features to automatically identify those contributing to improving the segmentation while discarding compromising ones. Consequently, the model can modify vessel intensities or textures while preserving their spatial arrangement and geometrical properties. The resulting outputs are thus hybrids between $\mathcal{S}$ and $\mathcal{T}$, intended to facilitate the segmentation process. This automatic alignment of the two domains allows us to discard data harmonization during pre-processing, which is often a domain-specific and time-consuming task.

\begin{figure*}[!t]
\centering
\includegraphics[width=\textwidth]{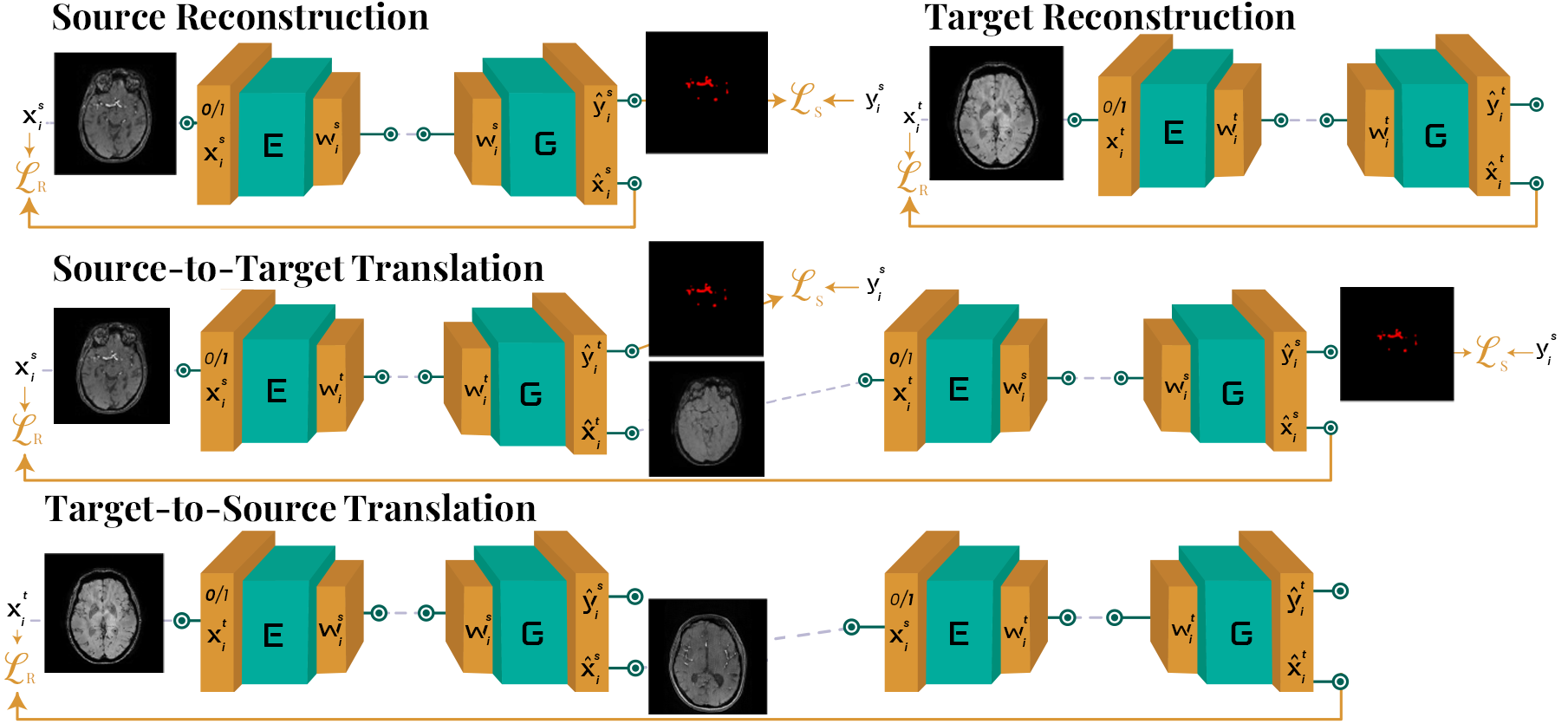}
\caption{In Phase 2 of our training algorithm, we perform both source and target reconstructions (first row, source domain on the left and target domain on the right) 
and source-to-target and target-to-source translations (second and third rows). 
The backpropagation of $\mathcal{L}_{\textrm{\textsc{r}}}$ exclusively updates the weights of $E$, while $\mathcal{L}_{\textrm{\textsc{s}}}$ influences both $E$ and $G$.} \label{Fig:supplementary}
\end{figure*}

\subsection{Cycle Consistency}
Up to this point, we have described the forward pass carried out during Phase 2 by $E$ and $G$ from a source image $x_i^{s}$ to its source reconstruction $\hat{x}_i^{s}$ and source-to-target translation $\hat{x}_i^{t}$, with corresponding predictions $\hat{y}_i^{s}$ and $\hat{y}_i^{t}$. This data flow remains equivalent when working with an input image $x_i^{t}$ from the target domain, resulting in one target reconstruction and one target-to-source translation. However, segmentation losses are only computed when annotations are available, i.e., in $T_L$. In both domains, feeding the model with opposite values of $d$ in succession corresponds to performing two complementary translations that neutralize each other's effects, as the first changes the input image's domain, while the second brings it back to its original one. This cyclical behavior can be exploited to enable the computation of cycle-consistency reconstruction losses alongside the intra-domain reconstruction losses, i.e., each translation is immediately followed by its inverse to enforce image fidelity. 
\revision{This approach reduces architectural complexity compared to traditional cycle-based methods, which typically necessitate two encoder-decoder pairs}~\citep{CycleGAN}. However, both the generator and encoder must be used twice during each cycle. Consequently, when the cycle is completed, the losses are propagated into $G$ and $E$ only once, taking into account the most recent pass. 
\revision{We compute both intra-domain and cycle-consistency reconstruction losses~$\mathcal{L}_{\textrm{\textsc{r}}}$ as:

\begin{equation}
\mathcal{L}_{\textrm{\textsc{r}}} = \mathcal{L}_{\textrm{MSE}} + \mathcal{L}_{\textrm{LPIPS}},
\end{equation}
where $\mathcal{L}_{\textrm{MSE}}$ is the mean squared error loss calculated between the input and the reconstructed image, and $\mathcal{L}_{\textrm{LPIPS}}$ is the perceptual similarity metric that measures differences in feature space as defined in~\cite{LPIPS}.}

\revision{Figure~\ref{Fig:supplementary} illustrates the steps involved in Phase 2. 
Moreover, Algorithm~\ref{alg:training} provides a formal description of the complete procedure, complementing the visual representation in Figures~\ref{Fig:sketch} and \ref{Fig:supplementary}.}
\begin{algorithm}[!ht]  
\SetAlgoLined
\caption{\revision{Two-Phase Training for Multi-Domain Vessel Segmentation}}
\label{alg:training}
\KwIn{Source dataset $S = \{(x_i^s, y_i^s)\}_{i=1}^N$, target datasets $T_U = \{x_i^t\}_{i=1}^M$, $T_L = \{(x_i^{lt}, y_i^{lt})\}_{i=1}^m$}
\KwOut{Trained generator $G$ and encoder $E$ for vessel segmentation}

\textbf{Phase 1:} Disentangled latent space learning\;
Initialize generator $G$ and discriminator $D$\;
Train $G$ and $D$ adversarially for 250,000 iterations using $\mathcal{L}_{\textrm{tot}} = \mathcal{L}_{\textrm{adv}}(G,D) + \mathcal{L}_{R_1}(D) + \mathcal{L}_{\textrm{pl}}(G)$\;

\textbf{Phase 2:} Label-preserving image-to-image translation\;
Initialize encoder $E$ and freeze parameters of $G$ except for label-synthesis branch ($G_{lsb})$\;
\For{iteration = 1 to 20000}{
    Sample batch from $S \cup T_U \cup T_L$\;
    
    \For{each source image $x_i^s \in S$}{
        // Source reconstruction ($d=0$) and source-to-target translation ($d=1$)\;
        $w_i^s \leftarrow E(x_i^s | d=0)$, $w_i^t \leftarrow E(x_i^s | d=1)$\;
        $\hat{x}_i^s \leftarrow G(w_i^s)$, $\hat{y}_i^s \leftarrow$ $G_{lsb}(w_i^s)$\;
        $\hat{x}_i^t \leftarrow G(w_i^t)$, $\hat{y}_i^t \leftarrow$ $G_{lsb}(w_i^t)$ \;
        Compute $\mathcal{L}_{\textrm{\textsc{r}}}(x_i^s, \hat{x}_i^s)$ and $\mathcal{L}_{\textrm{\textsc{s}}}(y_i^s, \hat{y}_i^s, \hat{y}_i^t)$\;
    }
    
    \For{each target image $x_i^t \in T_U \cup T_L$}{
        // Target reconstruction ($d=1$) and target-to-source translation ($d=0$)\;
        $w_i^t \leftarrow E(x_i^t | d=1)$, $w_i^s \leftarrow E(x_i^t | d=0)$\;
        $\hat{x}_i^t \leftarrow G(w_i^t)$, $\hat{y}_i^t \leftarrow$ $G_{lsb}(w_i^t)$ \;
        $\hat{x}_i^s \leftarrow G(w_i^s)$, $\hat{y}_i^s \leftarrow$ $G_{lsb}(w_i^s)$\;
        Compute $\mathcal{L}_{\textrm{\textsc{r}}}(x_i^t, \hat{x}_i^t)$\;
        \If{$x_i^t \in T_L$}{
            Compute $\mathcal{L}_{\textrm{\textsc{s}}}(y_i^{lt}, \hat{y}_i^t, \hat{y}_i^s)$\;
        }
    }
    
    Update $E$ by gradient descent on $\mathcal{L}_{\textrm{\textsc{r}}}$ and $\mathcal{L}_{\textrm{\textsc{s}}}$\;
    Update $G_{lsb}$ by gradient descent on $\mathcal{L}_{\textrm{\textsc{s}}}$\;
}
\Return{trained $G$ and $E$}
\end{algorithm}
\subsection{Inference}
Given a new image $x^t_{\text{new}}$, the model generates its reconstruction in $\mathcal{T}$, i.e., $\hat{x}^t_{\text{new}}$, and its translation in $\mathcal{S}$, i.e., $\hat{x}^s_{\text{new}}$. Simultaneously, $G_{lsb}$ retrieves the segmentation masks $\hat{y}^t_{\text{new}}$ and $\hat{y}^s_{\text{new}}$, corresponding respectively to $\hat{x}^t_{\text{new}}$ and $\hat{x}^s_{\text{new}}$. Since both predictions contain valuable information about vessel segmentation, the final segmentation mask is obtained by averaging $\hat{y}^t_{\text{new}}$ and $\hat{y}^s_{\text{new}}$ before the last argmax operation. Notably, the model only performs reconstruction when used with a source image $x^s_{\text{new}}$. The translation capability, which involves generating $\hat{x}^t_{\text{new}}$ and the corresponding $\hat{y}^t_{\text{new}}$, is not used, since our main goal is the segmentation of the target domain.

\section{Experiments and Results}
\label{sec:experiments}

\subsection{Experimental Setup}
In this section, we detail our experimental methodology for evaluating vessel segmentation across diverse medical imaging modalities. We first describe the datasets used in our experiments, including their characteristics and the preprocessing steps employed. We then provide implementation details of our framework, highlighting the network architectures and training procedures. Finally, we outline the evaluation metrics used.
\subsubsection{Datasets}
Our experiments use the following datasets. 

\noindent
\textbf{OASIS-3~\citep{OASIS3}}. We randomly select 49 time-of-flight (TOF) MRA volumes. These volumes have a median grid size of $576 \times 768 \times 232$ voxels and a median voxel size of $0.30 \times 0.30 \times 0.60$ mm. Our selection encompasses 27 cognitively normal subjects and 10 patients at different stages of cognitive decline, all adults ranging in age from 42 to 95 years. 


\noindent
\textbf{IXI\footnote{\url{https://brain-development.org/ixi-dataset}}}.  We sample 50 TOF MRA volumes, with a median grid size of $359 \times 481 \times 100$ voxels and a median voxel size of $0.47 \times 0.47 \times 0.80$ mm. All images were acquired from healthy subjects spanning an age range of 20 to 86 years. 

\noindent
\textbf{TopCoW~\citep{TopCow}}. We use the 40 CTA volumes within the first release of the dataset. The volumes exhibit a median grid size of $290 \times 366 \times 211$ voxels and a median voxel size of $0.46 \times 0.46 \times 0.70$ mm. The patients within this cohort were all in the process of recovering from disorders related to strokes. 

\noindent
\textbf{Susceptibility-weighted images (SWI).} We use a private dataset consisting of 28 SWI venographies from retrospective studies previously conducted at UCL Queen Square Institute of Neurology, Queen Square MS Centre, University College London. The images have a median grid size of $480 \times 480 \times 288$ voxels and a median voxel size of $0.50 \times 0.50 \times 0.50$ mm and include adult subjects showing no visible lesions on SWI. 


For IXI, we use the vessel annotations provided in \cite{falcetta2025vesselverse}.
For OASIS-3 and SWI, all image volumes were annotated by two experts (RC, MAZ) to obtain vessel masks. For TopCoW, we used the masks included in the dataset, which include annotations only of the vessels constituting the circle of Willis (CoW). Brain masks were obtained using SynthStrip~\citep{SynthStrip}. For TopCoW, we generated brain annotations through a registration and resampling procedure initiated from the pairwise MRA.

The datasets undergo separate pre-processing without any inter-domain harmonization. First, all volumes are resampled using bicubic interpolation to fix a uniform spacing, calculated as the dataset's median value, with minor increments made if the images do not fit into a volume of $512^3$ voxels. Next, each volume is rescaled based on its mean and standard deviation, and then clipped between the 0.1 and 99.9 percentiles and normalized in the range $[-1,+1]$. The segmentation masks undergo one-hot encoding, 
resulting in a three-dimensional label: one dimension for the brain, one for the vessels, and an additional one for the background. 

\subsubsection{Implementation Details}
Our framework is implemented in PyTorch 1.9.1. Phase 1 and Phase 2 use batches of four images each, and run for 250k and 20k iterations, respectively. In addition, a preliminary phase of 15k iterations is conducted before Phase 2, to pre-train the model using only source data. After training, the models with the best validation performance on $\mathcal{S}$ and $\mathcal{T}$ are selected for the final evaluation. The generator $G$ and discriminator $D$ are based on StyleGAN2~\citep{StyleGAN2}, while the label-synthesis branch is adapted from DatasetGAN~\citep{DatasetGAN}. As in~\citep{pixel2style2pixel}, the encoder $E$ maps input images into the extended latent space $\mathcal{W}+$ of StyleGAN2, using a ResNet backbone inspired by~\cite{Residuals}. Building upon this backbone, multiple outputs are branched out: one for latent code prediction and the other for feature tensor prediction. These branches are then connected to $G$ through a dynamic skip connection module~\cite{Yang2022}, which filters the residual information to establish fine-level content correspondences.
\revision{Our framework is configured for multi-class segmentation to extract both brain tissue and vessels simultaneously. The segmentation masks include three classes: background, brain, and vessels. 
}
All code and experiments can be accessed on \href{https://github.com/i-vesseg/MultiVesSeg}{github.com/i-vesseg/MultiVesSeg}.

\subsubsection{Evaluation setup}
We evaluate models based on their segmentation performance on the target datasets using test splits. We assess performance using the Dice coefficient (Dice), the centerlineDice (clDice)~\citep{Shit2021}, \revision{Average Symmetric Surface Distance (ASSD)}~\citep{ASSD}, precision, and recall. 

\subsection{Ablation Studies}
We study how the performance of our model is impacted by the number of available annotated images in both the target and source domains, as well as by various architectural choices. Given the substantial domain gap between angiographies and venographies, which depict two different vessel types, we utilize TOF MRA images from OASIS-3 as the source domain $\mathcal{S}$ and SWI images as the target domain $\mathcal{T}$ to analyze the behavior of our model in this particularly complex scenario.

\begin{table}[t]
\centering
\caption{Source domain performance on OASIS-3}
\label{Tab:DSC source}
\renewcommand{\arraystretch}{1.2}
\footnotesize
\begin{tabular*}{\columnwidth}{@{\extracolsep{\fill}}lcccc}
\bottomrule
 & Dice & Precision & Recall & clDice\\
\hline
Vessels & 73.7 ± 2.8 & 66.9 ± 4.8 & 82.5 ± 3.7 & 76.9 ± 5.2\\
\toprule
\end{tabular*}
\end{table}

\subsubsection{Intra-domain Performance}\label{sec:intra_domain}
We first assess the performance of our method in intra-domain vessel segmentation. In Phase 1, we include a source dataset ($S$) of $N=35$ source volumes and a target dataset ($T$) of $|T|=M+m=20$ target volumes into our training set. As this phase is entirely unsupervised, the division between the unlabeled and labeled target sets $T_{U}$ and $T_{L}$ does not have any impact. Subsequently, we pre-train the encoder $E$ and the segmentation branch of $G$ using only the source data (left half of the first row in Figure~\ref{Fig:supplementary}), ignoring source-to-target translation. For evaluation, we split equally the remaining 14 TOF MRAs between validation and testing, following a 70-15-15 ratio. The results on the testing set are presented in Table~\ref{Tab:DSC source}, demonstrating that our method's performance is comparable to state-of-the-art approaches for brain artery segmentation~\citep{Livne2019,Dang2022}.

\subsubsection{Impact of Target Annotations}\label{sec:target_test}
We investigate the model's sensitivity to the number of annotated target images. Using a fixed number of source images ($N=35$), we gradually increase the number of annotated target images into the training set ($T_L$). We begin with $m=0$ and progress to $m=1$ and $m=3$ midpoint slices, extracted from three distinct volumes. This sequence concludes with the inclusion of the full three volumes into $T_L$. The remaining volumes are used without annotations ($M=17$). Four images are set aside for validation, and another four are kept for testing.

\begin{figure*}
\centering
\includegraphics[width=\textwidth]{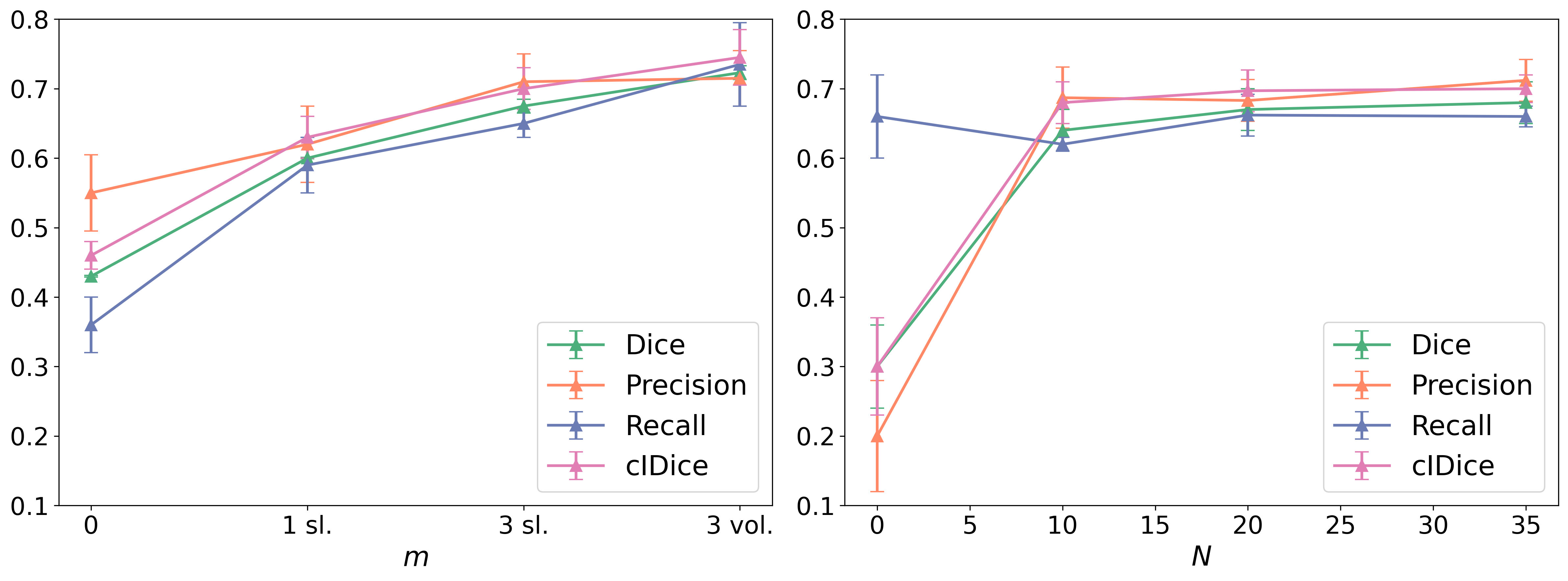}
\caption{\revision{Vessel segmentation performance with varying target annotations $m$ (left) and source annotations $N$ (right). Vertical error bars represent the standard deviation across the testing set.}} \label{Fig:plot_m_N}
\end{figure*}

Figure~\ref{Fig:plot_m_N} (left) reports vessel segmentation performance. As expected, the performance improves as the number of available annotated samples increases. In particular, there is a performance boost observed during the transition from $m=0$ to $m=1$ slice, marking the shift from an unsupervised DA scenario to a semi-supervised one. However, as the number of labeled slices increases from $m=3$ to cover three whole volumes ($m=831$ slices in total), the extent of this performance improvement gradually diminishes, suggesting a trend toward saturation. This indicates that while the model benefits from additional annotated target images, it already exhibits good behavior when only a few target labels are available. \revision{Based on this finding and the trade-off between annotation effort and performance, we set $m=3$ for all subsequent experiments in this study.}

\subsubsection{Impact of Source Annotations}
We investigate the scenario where the number of available source images varies ($N=[0,10,20,35]$) while the number of annotated target images is fixed ($m=3$ slices). Source and target validation and testing sets are the same as in Sections~\ref{sec:intra_domain} and ~\ref{sec:target_test}. Figure~\ref{Fig:plot_m_N} (right) summarizes the obtained results. We start considering a few-shot segmentation scenario, where minimal annotations are employed to train segmentation in $\mathcal{T}$, and there is no contribution from $\mathcal{S}$ (i.e., $N=0$). In this case, our model performs exclusively target reconstruction (right half of the first row in Figure~\ref{Fig:supplementary}), calculating the segmentation loss only for $T_{L}$. A sharp performance increase is observed when passing from $N=0$ to $N=10$ source volumes, i.e. when we activate source reconstruction and inter-domain translations. After, there is a modest increase of only $3.4\%$ in Dice when moving from $N=10$ to $N=35$. Despite the improvements becoming more gradual, overall, the contribution from the labeled source proves to be crucial for achieving satisfactory results. \revision{To maximize segmentation performance and make full use of our available data resources, we employed all $N=35$ source volumes for our subsequent experiments.}

\subsubsection{Architecture Elements}
We perform an ablation study to assess how the different elements in our architecture affect the segmentation accuracy. Specifically, we examine the effects of the following features, which we notice to have the most significant impact on the results: residual connections (Res.), to enhance information flow between $E$ and $G$; domain-specific batch normalization (DSBN), to normalize feature maps separately for the two domains; balanced data sampling (BDS), to ensure that each batch contains two samples from $S$, one from ${T}_L$ and one from ${T}_U$; and intensity inversion (Inv.), to flip the intensity values of the input images, thus mitigating the disparity between domains capturing vessels in dark and bright appearances respectively.

\begin{table}[t]
\caption{Architectural Choices}
\label{tab:Architectural Choices}
\renewcommand{\arraystretch}{1.2}
\begin{tabular*}{\columnwidth}{@{\extracolsep{\fill}}cccccc}
\toprule
\revision{Res.} & DSBN & BDS & \revision{Inv.} & Dice & clDice\\
\hline
\cmark & \cmark & \cmark & \cmark & \textbf{72.2 ± 2.5} & \textbf{75.4 ± 3.3}\\
\hline
\xmark & \cmark & \cmark & \cmark & 14.4 ± 3.4 & 17.0 ± 3.4\\
\hline
\cmark & \xmark & \cmark & \cmark & 69.3 ± 2.8 & 73.7 ± 3.2\\
\hline
\cmark & \cmark & \xmark & \cmark & 71.2 ± 2.4 & 74.4 ± 3.3\\
\hline
\cmark & \cmark & \cmark & \xmark & 71.8 ± 3.0 & 74.3 ± 3.3\\
\bottomrule
\end{tabular*}
\end{table}

Table~\ref{tab:Architectural Choices} displays the configurations obtained by deactivating each assessed component. Residual connections appear to exert the most influence on the model's functioning, causing a substantial drop in Dice from $72.2\%$ to $14.4\%$. Residual connections emerge as indispensable components, serving to preserve spatial information during reconstruction and facilitating the network's manipulation of low-level semantic attributes~\citep{Residuals}. Domain-specific batch normalization causes a drop of $2.9\%$; balanced data sampling $1.1\%$ and intensity inversion brings a negligible effect of $0.4\%$ in the Dice. Notably, intensity inversion is specific to MRA-to-MRV, thereby falling within the definition of data harmonization between the source and target domains. Proving that this inversion does not impact the performance reinforces the hypothesis that our method does not necessitate domain-specific pre-processing to address the domain gap. However, this is true only in the semi-supervised setting: after conducting an additional experiment with $m=0$, we notice a significant Dice score drop from $40.9\%$ to $0.1\%$ when intensity inversion is not used. This underlines the need for some form of guidance in establishing connections between vessels across TOF MRA and SWI modalities. This guidance could come in the form of labeled examples or intensity harmonization, but it represents an essential requirement for the correct functioning of our model.



\subsection{Comparison with State-of-the-Art Methods}\label{subsec:SOTACOMP}

\begin{table*}[ht]
\begin{minipage}{\textwidth}
\centering
\caption{Segmentation performance of different methods in the target domains. \revision{UniverSeg (both tasks) and SynthSeg (brain only) use the original pre-trained models.} }
\label{Tab:DSC}
\renewcommand{\arraystretch}{1.2}

\footnotesize
\resizebox{\textwidth}{!}{
\begin{tabular}{@{\extracolsep{\fill}}llccccccccc}
\multicolumn{11}{c}{\textbf{MC MRA}}\\
\bottomrule
 && U-Net & CycleGAN & SIFA & SynthSeg & UniverSeg$^*$ & AADG & DCDA & CS-CADA & Ours\\
\hline
\multirow{2}{*}{Dice} & Vessels & 65.6 ± 4.4 & 30.5 ± 3.3 & 53.7 ± 1.9 & 41.7 ± 5.4 & 7.3 ± 2.6 & 45.3 ± 2.9 & 12.0 ± 2.8 & 43.3 ± 5.0 & \textbf{69.9 ± 2.3}\\
& Brain & 95.1 ± 1.6 & 81.7 ± 2.9 & 89.4 ± 4.0 & 93.6 ± 3.3$^*$ & 95.4 ± 1.2 & 97.7 ± 0.3 & 90.5 ± 1.2 & 69.3 ± 6.3 & \textbf{97.8 ± 0.2}\\
\hline
\multirow{2}{*}{Precision} & Vessels & 62.6 ± 8.2 & 35.3 ± 5.1 & 56.5 ± 4.1 & 61.5 ± 7.4 & 5.4 ± 3.1 & \textbf{78.2 ± 4.1} & 23.1 ± 4.8 & 54.7 ± 15.1 & 70.0 ± 4.0\\
& Brain & 92.5 ± 3.0 & 86.8 ± 1.7 & 89.6 ± 1.0 & \textbf{99.2 ± 0.4$^*$} & 93.9 ± 2.3 & 98.8 ± 0.4 & 87.3 ± 1.0 & 81.7 ± 7.1 & 98.0 ± 0.5\\
\hline
\multirow{2}{*}{Recall} & Vessels & \textbf{70.3 ± 6.7} & 27.4 ± 4.4 & 51.7 ± 3.6 & 31.7 ± 4.9 & 14.7 ± 4.9 & 32.0 ± 3.1 & 8.2 ± 2.1 & 37.7 ± 4.7 & 70.2 ± 4.7\\
& Brain & \textbf{98.0 ± 0.5} & 77.4 ± 5.3 & 89.5 ± 7.6 & 88.9 ± 5.9$^*$ & 97.0 ± 0.5 & 96.6 ± 0.7 & 94.1 ± 2.9 & 60.5 ± 6.8 & 97.7 ± 0.5\\
\hline
clDice & Vessels & 68.7 ± 6.5 & 25.4 ± 3.0 & 51.5 ± 3.0 & 41.0 ± 6.4 & 8.1 ± 2.1 & 35.8 ± 2.6 & 8.4 ± 2.3 & 40.2 ± 5.4 & \textbf{76.8 ± 2.9}\\
\hline
\revision{ASSD} & Vessels & 3.39 ± 1.48 & 10.18 ± 3.36 & 3.38 ± 1.44 & 4.17 ± 0.86 & 15.59 ± 1.91 & 4.49 ± 1.42 & 11.59 ± 2.95 & 6.10 ± 1.90 & \textbf{1.55 ± 0.57}\\
\toprule
\end{tabular}
}

\resizebox{\textwidth}{!}{
\begin{tabular}{@{\extracolsep{\fill}}llccccccccc}
\multicolumn{11}{c}{\textbf{MRA-to-CTA}}\\
\bottomrule
 && U-Net & CycleGAN & SIFA & SynthSeg & UniverSeg$^*$ & AADG & DCDA & CS-CADA & Ours\\
\hline
\multirow{2}{*}{Dice} & Vessels & 70.5 ± 3.0 & 33.1 ± 4.3 & 60.9 ± 2.9 & 55.1 ± 23.6 & 11.5 ± 9.0 & 5.8 ± 6.0 & 0.0 ± 0.0 & 0.0 ± 0.0 & \textbf{74.5 ± 4.2}\\
& Brain & 95.8 ± 1.7 & 93.1 ± 1.6 & 94.4 ± 1.9 & 5.1 ± 7.0$^*$ & 95.9 ± 0.8 & 94.8 ± 1.6 & 91.2 ± 3.5 & 85.6 ± 10.4 & \textbf{96.6 ± 1.1}\\
\hline
\multirow{2}{*}{Precision} & Vessels & 72.7 ± 13.5 & 27.1 ± 5.2 & 63.5 ± 8.3 & 52.2 ± 15.6 & 38.9 ± 19.9 & 22.5 ± 22.3 & 0.0 ± 0.0 & 0.0 ± 0.0 & \textbf{73.3 ± 13.3}\\
& Brain & 94.2 ± 3.1 & 94.9 ± 1.4 & 94.1 ± 3.3 & 49.1 ± 49.1$^*$ & 93.5 ± 1.5 & 91.5 ± 3.4 & 95.1 ± 1.1 & 95.2 ± 0.9 & \textbf{96.2 ± 1.8}\\
\hline
\multirow{2}{*}{Recall} & Vessels & 72.1 ± 10.5 & 43.1 ± 2.5 & 59.1 ± 2.6 & 63.2 ± 27.7 & 6.9 ± 5.7 & 3.4 ± 3.5 & 0.0 ± 0.0 & 0.0 ± 0.0 & \textbf{78.8 ± 8.5}\\
& Brain & 97.5 ± 0.4 & 91.4 ± 2.2 & 94.8 ± 2.0 & 2.8 ± 3.8$^*$ & 98.5 ± 1.2 & \textbf{98.5 ± 1.5} & 87.8 ± 6.3 & 79.2 ± 15.2 & 97.1 ± 1.1\\
\hline
clDice & Vessels & 72.5 ± 6.7 & 39.7 ± 5.2 & 68.9 ± 6.3 & 63.5 ± 29.0 & 18.0 ± 9.0 & nan ± nan & nan ± nan & nan ± nan & \textbf{78.0 ± 8.7}\\
\hline
\revision{ASSD} & Vessels & 2.40 ± 0.99 & 3.64 ± 0.53 & 2.12 ± 0.56 & 4.01 ± 5.82 & 12.24 ± 3.10 & 13.92 ± 6.99 & 12.97 ± 2.11 & nan ± nan & \textbf{1.46 ± 0.49}\\
\hline
\toprule
\end{tabular}
}

\resizebox{\textwidth}{!}{
\begin{tabular}{@{\extracolsep{\fill}}llccccccccc}
\multicolumn{11}{c}{\textbf{MRA-to-MRV}}\\
\bottomrule
 && U-Net & CycleGAN & SIFA & SynthSeg & UniverSeg$^*$ & AADG & DCDA & CS-CADA & Ours\\
\hline
\multirow{2}{*}{Dice} & Vessels & 29.1 ± 4.9 & 5.1 ± 0.3 & 0.8 ± 0.5 & 10.9 ± 1.9 & 3.6 ± 1.1 & 2.0 ± 1.2 & 0.0 ± 0.0 & 0.4 ± 0.2 & \textbf{67.5 ± 1.7}\\
& Brain & 83.0 ± 2.5 & 75.0 ± 0.8 & 91.4 ± 1.9 & 97.4 ± 0.1$^*$ & 83.5 ± 2.5 & 96.7 ± 0.5 & 75.6 ± 1.1 & 25.7 ± 2.5 & \textbf{97.8 ± 0.2}\\
\hline
\multirow{2}{*}{Precision} & Vessels & 18.2 ± 4.0 & 11.5 ± 0.8 & 2.6 ± 1.1 & 51.4 ± 5.7 & 6.5 ± 2.4 & 1.3 ± 0.6 & 2.4 ± 2.1 & 0.8 ± 0.4 & \textbf{71.1 ± 4.4}\\
& Brain & 71.3 ± 3.7 & 62.6 ± 0.7 & 97.8 ± 0.2 & 96.9 ± 0.5$^*$ & 72.8 ± 3.8 & 97.4 ± 0.3 & 62.0 ± 1.6 & 32.5 ± 3.8 & \textbf{97.8 ± 0.4}\\
\hline
\multirow{2}{*}{Recall} & Vessels & \textbf{76.0 ± 5.5} & 3.3 ± 0.3 & 0.5 ± 0.3 & 6.1 ± 1.2 & 2.5 ± 0.7 & 6.3 ± 4.0 & 0.0 ± 0.0 & 0.3 ± 0.1 & 64.4 ± 2.1\\
& Brain & \textbf{99.5 ± 0.2} & 93.6 ± 1.2 & 85.9 ± 3.5 & 97.9 ± 0.5$^*$ & 98.1 ± 0.2 & 96.0 ± 0.9 & 97.0 ± 0.5 & 21.3 ± 1.8 & 97.8 ± 0.5\\
\hline
clDice & Vessels & 33.5 ± 5.9 & 4.1 ± 0.3 & 0.6 ± 0.4 & 10.4 ± 1.9 & 2.7 ± 0.9 & 1.8 ± 1.1 & nan ± nan & 0.4 ± 0.2 & \textbf{69.9 ± 2.7}\\
\hline
\revision{ASSD} & Vessels & 9.30 ± 0.53 & 10.54 ± 0.29 & 15.04 ± 6.98 & 14.41 ± 1.41 & 20.32 ± 2.52 & 10.46 ± 6.46 & 60.66 ± 5.62 & 18.73 ± 0.96 & \textbf{1.42 ± 0.20}\\
\hline
\toprule
\end{tabular}
}

\end{minipage}
\end{table*}

We compare the best results obtained through our ablation studies with \revision{a 2D U-Net~\citep{UNet} as a baseline and} seven DA state-of-the-art methods. 
\revision{For the baseline, we implement a fully-supervised approach with limited target annotations, training only on the few target samples in $T_L$ (m=3 slices to maintain methodological consistency). This implementation\footnote{https://github.com/wolny/pytorch-3dunet} uses a single input channel and three output classes (background, brain, vessels). The architecture consists of four encoding/decoding stages with feature maps ranging from 64 to 512. We optimize using Adam (lr=2e-4) with a combined BCE and Dice loss function, processing 512×512 2D slices and employing validation-based early stopping.} The seven DA SOTA methods are: CycleGAN~\citep{CycleGAN}, Synergistic Image and Feature Adaptation (SIFA)~\citep{SIFA}, SynthSeg~\citep{SynthSeg}, UniverSeg~\citep{UniverSeg}, Automatic Augmentation for Domain Generalization (AADG)~\citep{AADG}, DCDA~\citep{DCDA}, and Contrastive Semi-supervised learning for Cross Anatomy Domain Adaptation (CS-CADA)~\citep{CS-CADA}. In particular:

\noindent\textbf{1)} CycleGAN is a well-established method to perform unpaired image-to-image translation on natural images. After translating data from $\mathcal{T}$, we feed the results into a 2D U-Net previously trained on $S$, as CycleGAN does not provide segmentation.

\noindent\textbf{2)} SIFA is an UDA technique based on image-to-image translation for multi-class medical segmentation, therefore trained using both $S$ and $T$, without utilizing any target label $y_i^{lt}$;

\noindent\textbf{3)} SynthSeg is a UDA 3D output-level alignment method based on synthetic data generation for brain synthesis and segmentation. It is trained with masks $y_i^s$ from $S$, determining the best checkpoint based on the target performance;

\noindent\textbf{4)} UniverSeg is a foundation model focusing on unseen medical segmentation tasks without additional training;

\noindent\textbf{5)} AADG is a multi-source domain generalization framework based on data manipulation of retinal vessel images. To leverage training from multiple source domains, the network is trained using all datasets except the target;

\noindent\textbf{6 and 7)} DCDA and CS-CADA are, respectively, unsupervised and semi-supervised DA methods designed for retinal vessel segmentation and 2D coronary artery segmentation. We train these methods using both $S$ and $T$, including labels from $T_L$ for CS-CADA.

Using OASIS-3 as a source domain, we conduct experiments in three distinct domain adaptation scenarios of increasing difficulty to ensure a broader perspective, comparing our model's performance in adapting to the following shifts:

\noindent\textbf{1)  Multi-center (MC) MRA}, where MRAs are used as $\mathcal{S}$ and $\mathcal{T}$, but from different centers. Thirty-six unlabeled volumes ($T_U$) from IXI enter the training set; seven are kept for validation and seven for testing; \\
\noindent\textbf{2)  MRA-to-CTA}, where the target domain is CTAs from TopCoW, including 28 volumes without annotations ($T_U$) for training, six for validation and six for testing; and \\
\noindent\textbf{3)  MRA-to-MRV}, with SWIs used as $\mathcal{T}$, of which 20 volumes deprived of labels ($T_U$) are included in the training set, four in the validation set and four in the testing set.

In MC MRA and MRA-to-MRV, we extract three midpoint slices from $T_U$ to form $T_L$. In MRA-to-CTA, we allocate three entire volumes for $T_L$ since they only have CoW annotations. For the source dataset, 35 are allocated for training ($S$), while seven volumes are used for validation and testing. 

\begin{figure*}[t]
\includegraphics[width=\textwidth]{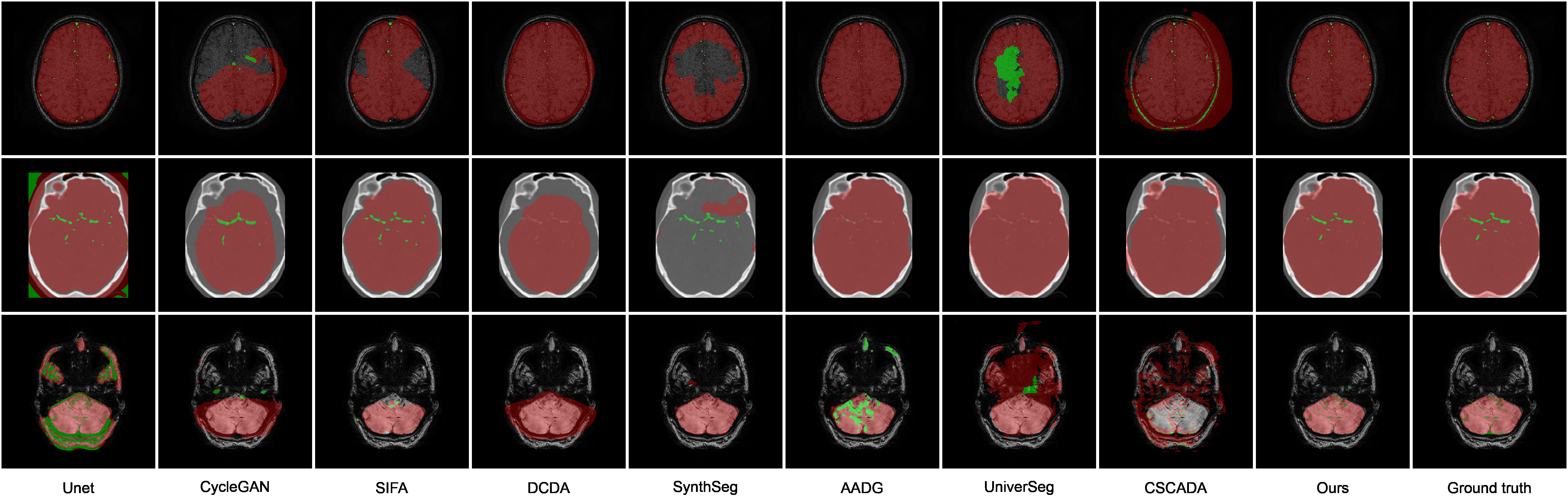}
\caption{Comparison of the segmentation results for brain and vessels in the target MRA, CTA, and SWI images using different methods. Red indicates brain masks, while green represents vessels. The rows display slices at varying levels: top, middle, and bottom.} \label{Fig:results}
\end{figure*}

\revision{For the sake of fairness, we evaluate both cross-modality vessel and brain segmentation since most of the methods (e.g., SIFA, SynthSeg, AADG, and UniverSeg) have been developed for segmenting large objects, such as the brain. To ensure consistent comparison, all methods were retrained for brain and vessel segmentation with two exceptions: UniverSeg was used as originally designed for zero-shot segmentation on unseen medical tasks without retraining, and SynthSeg utilized its original pre-trained model for brain tissue segmentation only, as it was specifically designed for this task. Notably, while the pre-trained SynthSeg model performs well with magnetic resonance images (both MRA and MRV), it fails to segment brain tissue in CTA images due to the fundamental differences in image contrast mechanisms.} Table~\ref{Tab:DSC} summarizes the obtained results, and Figure~\ref{Fig:results} displays a visual comparison of the results across MC MRA, MRA-to-CTA, and MRA-to-MRV.

In general, most methods have a very good performance on the brain task, but there is a clear difficulty in segmenting the vessels. %
This becomes particularly visible in the MRA-to-MRV scenario: both the U-Net baseline, trained with full supervision on the reduced dataset $T_L$, and the considered state-of-the-art methods in domain adaptation and generalization struggle to segment veins. \revision{The performance degradation trend from} MC MRA and MRA-to-CTA to MRA-to-MRV highlights the challenge posed by increasing domain gaps. UniverSeg fails to segment vessels across all scenarios but demonstrates satisfactory brain segmentation performance despite not requiring additional training. CS-CADA, the only other SSDA model besides ours, provides poor results overall, likely because it originally relies on a larger annotated target set than $T_L$.

Notably, our proposed method achieves high performance in the target domain for both brain and vessel segmentation. In particular, it bridges even the widest domain gaps, successfully segmenting veins using only three annotated target slices and leveraging information from the associated arteries in the source modality. This demonstrates the model's capability to cope with the differences between arteries and veins, which are not limited to their low-level attributes, such as intensities and textures, but also encompass higher-level aspects, including their location and shape.

Our better performance in cross-modality vessel segmentation, especially in comparison to other techniques following similar translation paradigms, can be explained by the key elements introduced in our framework. Compared to CycleGAN, which requires separate networks for bidirectional translations and performs image-to-image translation independent of segmentation, our approach integrates translation and segmentation through our label-synthesis branch, directly enforcing label preservation. 
Unlike SIFA, which aligns feature distributions without distinguishing between vessel-related and volume-related features, our disentanglement mechanism specifically isolates vessel properties that can be safely translated without compromising spatial information. 
While CS-CADA introduces contrastive learning for domain adaptation,  it cannot selectively translate domain-specific features that preserve vessel positions and morphology. Similarly, DCDA's disentangling approach focuses primarily on style transfer for retinal vessels, which is insufficient for handling the complex domain shifts between cerebrovascular structures with different anatomical characteristics. 
Finally, our two-phase training strategy isolates adversarial learning to only the initial phase, providing more stable convergence compared to methods like CycleGAN and SIFA which rely on continuous adversarial training throughout the adaptation process. 
These architectural choices collectively enable our framework to effectively bridge the challenging domain gaps in cerebrovascular segmentation while maintaining accurate vessel delineation.






\begin{figure*}[!t]
\includegraphics[width=\textwidth]{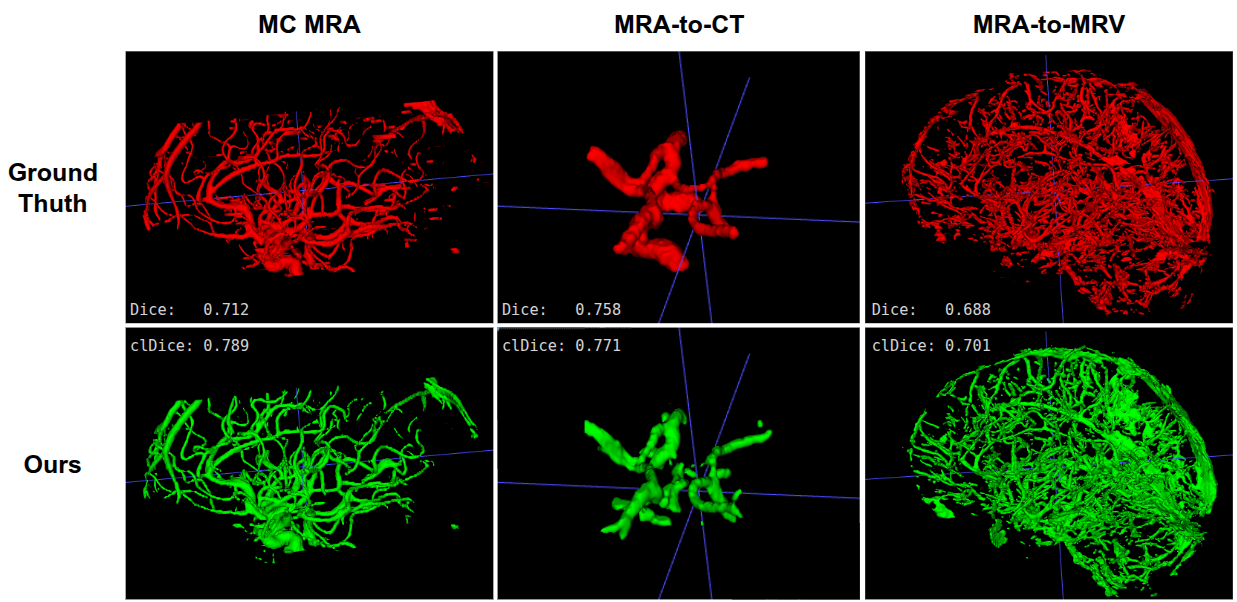}
\caption{\revision{3D visualization of vessel segmentation results across the three domain adaptation scenarios. Ground truth (red) and our method's predictions (green) are shown as 3D renderings with corresponding Dice and clDice scores.}} \label{Fig:3dresults}
\end{figure*}



\revision{Figure~\ref{Fig:3dresults} provides 3D visualizations of our segmentation results across the three domain adaptation scenarios, demonstrating the preservation of vessel topology and spatial continuity. The 3D renderings show that our method maintains the complex branching patterns and connectivity of the cerebrovascular tree across all tested scenarios. Notably, even in the most challenging MRA-to-MRV scenario, where the method must bridge the gap between arterial and venous structures with distinct anatomical characteristics, these visualizations reveal successful preservation of the overall vascular architecture with high fidelity to the ground truth topology.}

\subsection{\revision{Qualitative} Analysis}

\begin{figure*}[t]
\centering
\includegraphics[width=\textwidth]{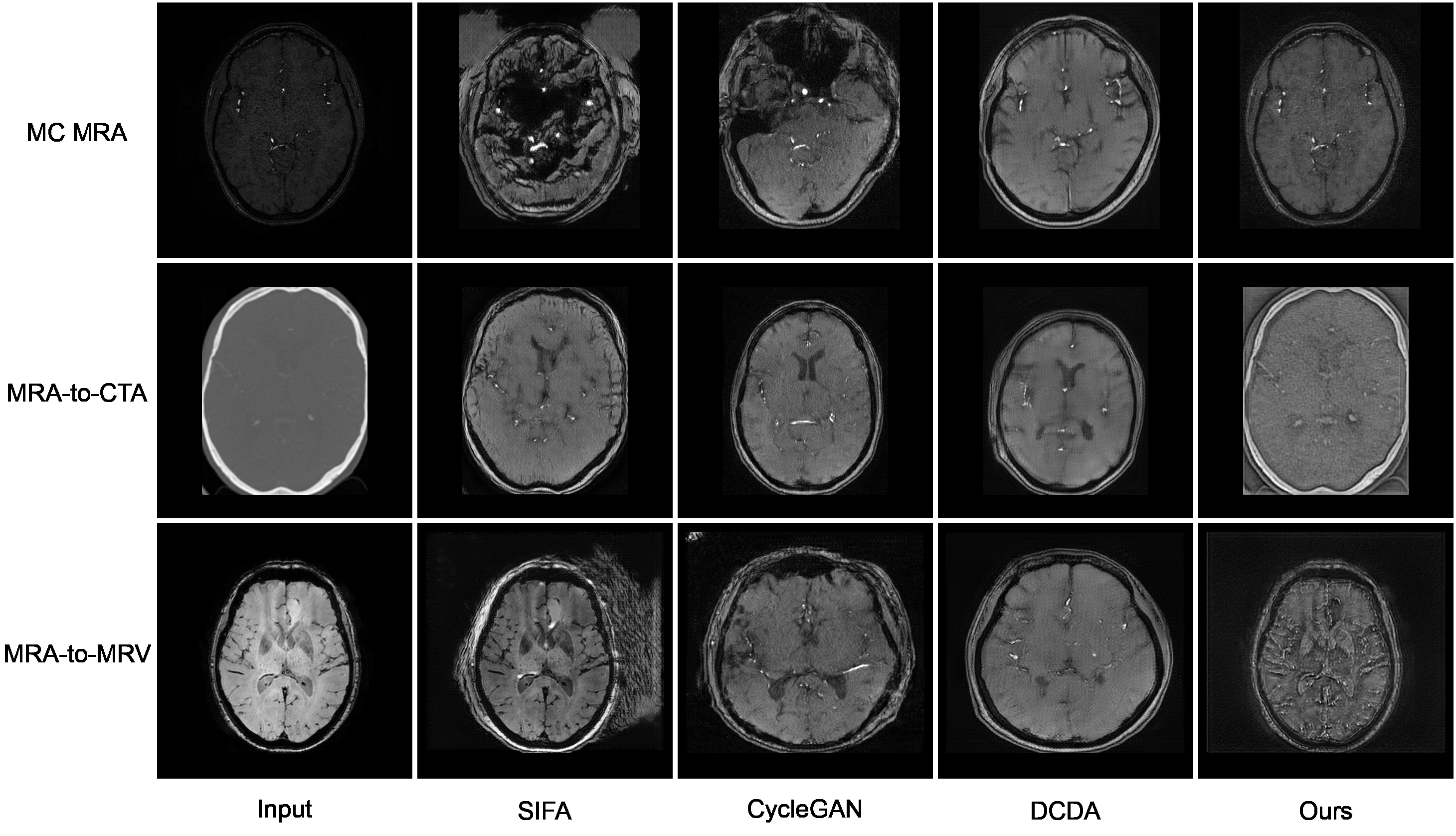}
\caption{Target-to-source translations produced by the different image-level alignment methods.}
\label{Fig:Inter-domain translations}
\end{figure*}

Adopting the path length regularization~\citep{StyleGAN2} has an important role in the domain adaptation process as it allows the disentangling of the latent space $\mathcal{W}$, enabling inter-domain translations that can handle independently volume-related image properties, such as overall spatial information and appearance, and vessel-related properties, such as their intensities, textures, shapes, locations, and densities. This allows preserving the target content while mimicking the appearance of a source image as it is better recognized by the segmentation branch. Keeping vessel position and shape unchanged, despite these being domain-specific features, is a key property to guarantee correct segmentation. By relying on the aforementioned capabilities, we have gathered evidence of the ability to separate the vessel-related features by visually inspecting the target-to-source translations generated by our model compared to other image-level alignment methods.


In Figure~\ref{Fig:Inter-domain translations}, we display three cases of translation: one from MC MRA (first row), another from MRA-to-CTA (second row), and a third from MRA-to-MRV (third row). These examples highlight how the different models act on the vessel-related properties. 
In particular, we identified three problematic behaviors that compromise the accuracy of the final segmentation results. These behaviors involve the translation of label-altering features due to their domain-specific nature. Firstly, vessels undergo displacement, resulting in changes to their position and size. This occurs while resizing the whole brain to align with the pixel spacing of the source domain. Specifically, CycleGAN and DCDA tend to translate all domain-specific features without distinction, including, in fact, the pixel spacing, and thus leading to spatial misalignment between the source and target domains. We believe this problem arises because the segmentation loss does not influence the prior translation enough, which is exactly the case in CycleGAN, where translation and segmentation are completely separate. Secondly, vessels are observed merging with the background and vanishing. This is noticeable as the number of bright vessels in the translations is never greater than the vessels in the target domain. The phenomenon is particularly evident in SWI images, where veins are generally more abundant than arteries in TOFs. The third issue arises in SIFA, which initially appears to better preserve the positions and shapes of the brain and vessels during translation, despite generating some shadow artifacts around the skull in MC MRA and MRA-to-MRV translations. However, most veins from SWIs are left untransformed and do not resemble arteries after translation. Only a few veins, likely those aligning well with the typical artery arrangement, transform into bright vessels. We attribute this behavior to the network's inability to link arteries and veins during translation without some form of guidance.


These findings align with what was observed for Figure~\ref{Fig:results}, where problematic vessels are either omitted from the final segmentation or displaced. Also, this reinforces the importance of enforcing label-preserving translations in our problem. Notably, our model uniquely transforms dark vessels from the input (SWI) into bright vessels without relocating them or reducing their number to replicate the typical arrangement of arteries in TOF MRA images. This ability to selectively translate only some domain-specific features, particularly those unrelated to vessel size and position, enables our approach to adapt veins and arteries and retrieve accurate segmentations. Lastly, we emphasize that achieving a hyper-realistic translation of target volumes is not the central focus of our model. We acknowledge that our translations may not appear entirely source-like but rather appear as hybrid representations. Indeed, the ability of the network to translate input images aims exclusively to serve the segmentation process, which is the primary objective of the proposed method.

\subsection{Computational Costs}
\revision{
We evaluate the computational efficiency of our approach. All experiments were conducted on two NVIDIA GeForce RTX 2080 Ti GPUs, each with 12GB of memory. For our framework, Phase 1 processes at approximately 0.42 iterations per second, requiring about 166 hours ($\approx$ 7 days) to complete 250,000 iterations. Phase 2 runs at approximately 0.20 iterations per second, taking about 28 hours to complete 20,000 iterations. This results in a total training time of approximately 194 hours ($\approx$ 8 days).

The computational cost of our approach stems primarily from the comprehensive disentanglement process in Phase 1, which builds a semantically rich latent space capable of independently representing various domain-specific features. This intensive investment enables our model to effectively handle challenging domain gaps without requiring domain-specific preprocessing steps that would otherwise add significant manual effort during data preparation. Additionally, Phase 2 involves complex image-to-image translations and cycle-consistency operations that contribute to the computational demands but are essential for preserving label integrity across domains.

Despite the high training costs, our framework offers a favorable trade-off between computational requirements and segmentation performance, particularly in challenging domain adaptation scenarios where other methods fail to produce acceptable results. The computational overhead can be justifiable when considering the performance improvements, especially when considering domains with a significant gap. 
}

\section{Discussion and Conclusion}
\label{sec:conclusion}
In this paper, we presented an end-to-end semi-supervised domain adaptation framework designed as an out-of-the-box tool for segmenting arteries and veins in images from different centers and/or modalities. To this end, we opted for a minimal pre-processing strategy that avoids any data harmonization between source and target domains. While enhancing the versatility of our model, this comes at the cost of widening the domain gap between the two domains. Our investigations analyzed this trade-off, delving into the concepts and mechanisms crucial for the effective functioning of our model. To address the problem of domain shift arising from different medical centers, imaging modalities, and vessel types, we rely on the path length regularization~\citep{StyleGAN2}, which allows for representing heterogeneous volumetric data in a unified and disentangled latent space. 
Consequently, we explored the potential of disentanglement, investigating the possibility of modifying selected domain-specific features to achieve inter-domain translation in a label-preserving manner.
\revision{One important design choice in our framework is the adoption of a 2.5D approach rather than full 3D processing. While our method targets 3D vessel segmentation, we process volumetric data as a series of multi-channel 2D slices. This technique enables larger batch sizes and reduces memory requirements, making the method more accessible with limited computational resources. We acknowledge that this approach may affect the continuity of vessels across distant slices. However, our evaluation using topology-aware metrics like clDice demonstrates that the method effectively preserves the topological structure of the cerebrovascular tree. This practical balance between accuracy and efficiency is particularly relevant for clinical applications where computational resources may be constrained.}
In addition to assessing the efficacy of disentanglement, we conducted ablation studies to determine the optimal number of source and target annotations and to evaluate the influence of key architectural choices on performance. Finally, we compared our framework against other state-of-the-art domain adaptation and domain generalization methods. Our approach demonstrates superior performance, accurately segmenting 3D brain vessels primarily using annotations from arterial images, which are comparatively easier to obtain. The results exhibit promising performance in semi-supervised domain adaptation scenarios, overcoming the difficulties posed by large domain gaps, in particular between veins and arteries, and the intricate morphology of the cerebrovascular tree. Despite our accomplishments, we acknowledge the potential for improvement. First, we highlight the necessity of our model to repeat training for each new target domain, and we note that in-context learning, as offered by methods like UniverSeg, presents a viable alternative. Furthermore, our model requires guidance in the form of $m$ target annotated 2D slices. Again, foundation models can prove beneficial: by pre-training on extensive collections of tree-like objects, segmentation models can acquire a broader representation of vessels. This approach facilitates linking vessels from distant modalities without relying on any additional guidance.

\newpage


\acks{This work has received funding from the French government under management of ANR as part of the “Investissements d’avenir” program, reference ANR-19-P3IA-000 (3IA Côte d’Azur) and reference ANR-19-
P3IA-0001 (PRAIRIE 3IA Institute). Ferran Prados receives funding from the National Institute for Health Research (NIHR), the Biomedical Research Centre initiative at University College London Hospitals (UCLH). Ferran
Prados received a Guarantors of Brain fellowship 2017-2020. Ninon Burgos has received funding from the ANR-10-IA Institut Hospitalo-Universitaire-6 (ANR-10-IAIHU-06). Maria A. Zuluaga and Daniele Falcetta are funded by an ANR
JCJC project, I-VESSEG  (22-CE45-0015-01).}

%
\ethics{The work follows appropriate ethical standards in conducting research and writing the manuscript, following all applicable laws and regulations regarding treatment of animals or human subjects.}


\coi{We declare we do not have any conflicts of interest}

\data{All datasets utilized in this study are publicly available and can be accessed freely. However, access to some of these datasets requires prior registration or application for access due to the sensitive nature of the data or to comply with data protection regulations. Despite the public availability of the datasets, direct sharing of the data by the authors is not permissible due to copyright and licensing restrictions imposed by the data providers. We encourage interested researchers to obtain the data directly from the respective repositories and websites, adhering to the specified access procedures and usage policies. Where applicable, the authors will share the labels obtained from their analyses.}

\bibliography{melba_accepted.bib}
\end{document}